\documentclass[conference]{IEEEtran}
\pdfoutput=1
\IEEEoverridecommandlockouts
\usepackage[hidelinks]{hyperref}
\usepackage{url}
\usepackage{cite}
\usepackage{amsmath,amssymb,amsfonts}
\usepackage{algorithmic}
\usepackage{graphicx}
\usepackage{textcomp}
\usepackage{xcolor}
\def\BibTeX{{\rm B\kern-.05em{\sc i\kern-.025em b}\kern-.08em
    T\kern-.1667em\lower.7ex\hbox{E}\kern-.125emX}}
\begin{document}

\title{AI-based BMI Inference from Facial Images: An Application to Weight Monitoring}

\author{\IEEEauthorblockN{Hera Siddiqui$^{1}$, Ajita Rattani$^{1}$, Dakshina Ranjan Kisku$^{2}$, Tanner Dean$^{3}$}
\IEEEauthorblockA{\textit{$^{1}$Wichita State University, Wichita, USA} \\
\textit{$^{2}$National Institute of Technology, Durgapur, India}\\
\textit{$^{3}$University of Kansas Medical Center, Kansas, USA}\\
hxsiddiqui@shockers.wichita.edu; ajita.rattani@wichita.edu; drkisku@cse.nitdgp.ac.in; tdean@kumc.edu}
}

\maketitle

\begin{abstract}
Self-diagnostic image-based methods for healthy weight monitoring is gaining increased interest following the alarming trend of obesity.
Only a handful of academic studies exist that investigate AI-based methods for Body Mass Index (BMI) inference from facial images as a solution to healthy weight monitoring and management.
To promote further research and development in this area, we evaluate and compare the performance of five different deep-learning based Convolutional Neural Network (CNN) architectures i.e., VGG19, ResNet50, DenseNet, MobileNet, and lightCNN for BMI inference from facial images. Experimental results on the three publicly available BMI annotated facial image datasets assembled from social media, namely, VisualBMI, VIP-Attributes, and Bollywood datasets, suggest the efficacy of the deep learning methods in BMI inference from face images with minimum Mean Absolute Error (MAE) of $1.04$ obtained using ResNet50.
\end{abstract}


%
\section{Introduction}
\label{sec:intro}
Describable visual attributes are any visual and contextual information that can be gleaned from images~\cite{6248021,selfie}. Typically, these attributes can be \emph{automatically} deduced from images and can be broadly classified into demographic, anthropometric, medical, material, and behavioral attributes. Examples of demographic attributes include age, gender, and ethnicity, which can be gleaned from facial images~\cite{bio_iet,8272766}. Anthropometric attributes include body geometry and facial geometry. Body Mass Index, wrinkles, and health conditions are examples of medical attributes, while eyeglasses, scarf, and gait are examples of material and behavioral attributes~\cite{jain04}. Attributes such as age and gender have drawn significant interest in applications such as surveillance, forensics, human-computer interaction, indexing, and targeted advertisement systems~\cite{8272766}.

Recently, body weight and BMI have drawn significant interest in health-monitoring and weight loss applications~\cite{Wen13,8546159,Kocabey}.
 BMI is defined as (body mass in kg)/ (body height in m)$^2$. The BMI within $25.0$ to $30$ and $30$ or higher, falls within the overweight and the obese range, respectively.
This is due to alarming trends related to obesity affecting $93.3$ million adults in the United States alone~\cite{hales}. Obesity is one of the biggest drivers of preventable chronic diseases and healthcare costs in the United States. Severe obesity costs the United States approximately $69$ billion overall, with almost $8$ billion a year being paid for via state Medicaid programs~\cite{Wang}.
Chronic conditions related to obesity include \textit{heart disease, stroke, type $2$ diabetes, and some cancers}, which are the leading causes of preventable death\footnote{https://www.cdc.gov/obesity/}. Since $1975$, the worldwide prevalence of obesity has nearly tripled\footnote{https://www.who.int/news-room/fact-sheets/detail/obesity-and-overweight}.

A goal of Healthy People $2020$\footnote{https://www.healthypeople.gov/} program in the United states is to promote health and reduce chronic disease by maintaining a healthy body weight. This is in order to minimize the chances of chronic disease development and death at an earlier age. The current trend is in the investigation of image-based automated self diagnostic methods\footnote{https://www.healio.com/endocrinology/obesity/news/online/\%7Bd61fdec1-dbbb-4b6f-98e8-825f6d19c9b4\%7D/mobile-apps-may-facilitate-weight-loss-among-adults-with-type-2-diabetes} for healthy weight monitoring. Specific interest is in the development of face-based \emph{non-intrusive health care/ telemedicine solutions} for smartphones~\cite{Mann2020}. This interest has been spurred with wide-scale integration of face recognition technology in smartphones for legitimate access to mobile users such as iPhone X series~\cite{selfie}.

A handful of academic studies~\cite{Wen13, Kocabey,8546159} suggest that BMI can be gleaned from facial images using machine and deep learning methods. \emph{However, current literature lacks in understanding of the efficacy of different deep learning architectures, such as VGG and ResNet, in BMI prediction from facial images}. There is no evaluation and understanding of the efficacy of various deep CNN architectures in BMI prediction. Different CNN architectures are bound to obtain different results due to feature representation differences emerging from their unique architectures. Therefore, the comparison of different CNNs become vital in further advancing knowledge in this domain.

\vspace{0.25 cm} \noindent\textbf{Our Contribution:}
To advance the state-of-the-art in face analysis based automated self-diagnostic methods for maintaining a healthy weight, the contributions of this paper are as follows:
\begin{itemize}
\item Investigation and comparative analysis of deep features from different CNN architectures for BMI prediction from facial images. To facilitate this, deep features extracted from VGG, ResNet, MobileNet, and DenseNet models, pre-trained on ImageNet dataset, are evaluated in this study.
\item Experimental investigation on three publicly available datasets consisting of facial images from Asian, Caucasian, and African American, including Hollywood and Bollywood celebrities, in social media. These datasets have significant variations in facial images due to factors such as make-up, pose, and varying lighting conditions.
\end{itemize}

\section{Prior Work on BMI Inference from Facial Images}
\label{sec:format}
Wen and Guo~\cite{Wen13} proposed the first study, to the best of our knowledge, on the automated face-based estimation of BMI. The authors used geometry, and ratio-based features (such as cheekbone to jaw width, width to upper facial height ratio,  perimeter to area ratio, and eye size)  obtained using an Active Shape Model along with the Support Vector Regression~(SVR) for BMI prediction. Reported results obtained Mean Absolute Error (MAE) in the range $[2.65, 4.29]$ on MORPH-II face dataset. The BMI-annotation of MORPH-II is not publicly available.

Kocabey et al.~\cite{Kocabey} proposed a BMI prediction method composed of deep feature extraction using VGG in combination with Support Vector Regression. Experiments on the VisualBMI dataset assembled by the authors from the web obtained Pearson correlation of $0.71$, $0.57$, and $0.65$, for Male, Female, and Overall, respectively.

Dantcheva et al.~\cite{8546159} proposed an end-to-end CNN obtained by replacing the last fully connected layer of ResNet from $1000$ channels to $1$ channel and using smooth L$1$ loss to cater regression. Experiments on the VIP attribute dataset consisting of Hollywood celebrities assembled by the authors from the web obtained MAE of $2.32$, $2.30$, and $2.36$ for Male, Female, and Overall, respectively.



\section{Deep Features for BMI Prediction from Facial Images}

\subsection{Deep features from Pretrained CNNs}
The deep learning models evaluated in this study are mostly CNNs pre-trained on large scale datasets comprising of million of training images for large-scale image classification. Deep features are extracted by activating one of the layers of the network and obtaining feature representation. Table~\ref{tab:model_comp} shows the feature size and the number of parameters of the various CNN models used for deep feature extraction in this study. Next, we discuss these CNN models in terms of their architecture, and parameters involved due to the convolutional and fully connected layers.

\begin{enumerate}

\item \textbf{VGG}: The VGG architecture was introduced by Visual Graphics Group research team at Oxford University~\cite{Simonyan14c}. The architecture consists of sequentially stacked $3\times 3$ convolutional layers with intermediate max-pooling layers followed by a couple of fully connected layers for feature extraction. Usually, VGG models have $13$ to $19$ layers. We used VGG-$19$ in this study which has $140M$ number of parameters. 

\begin{figure}[!t]
\centering
\includegraphics[scale=0.8]{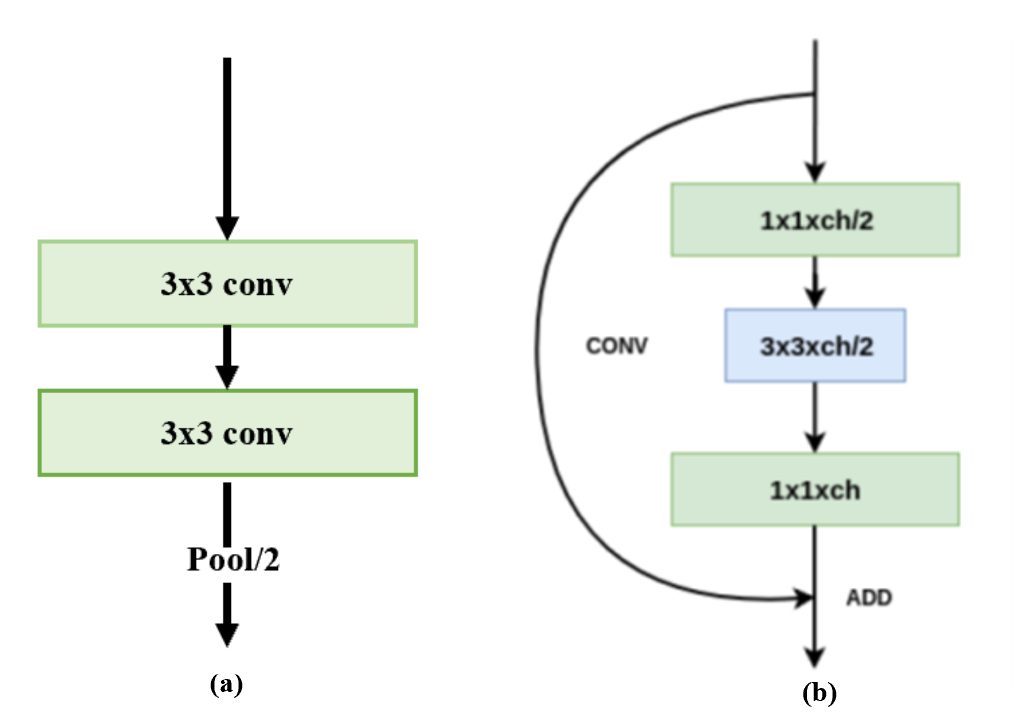}
\includegraphics[scale=0.8]{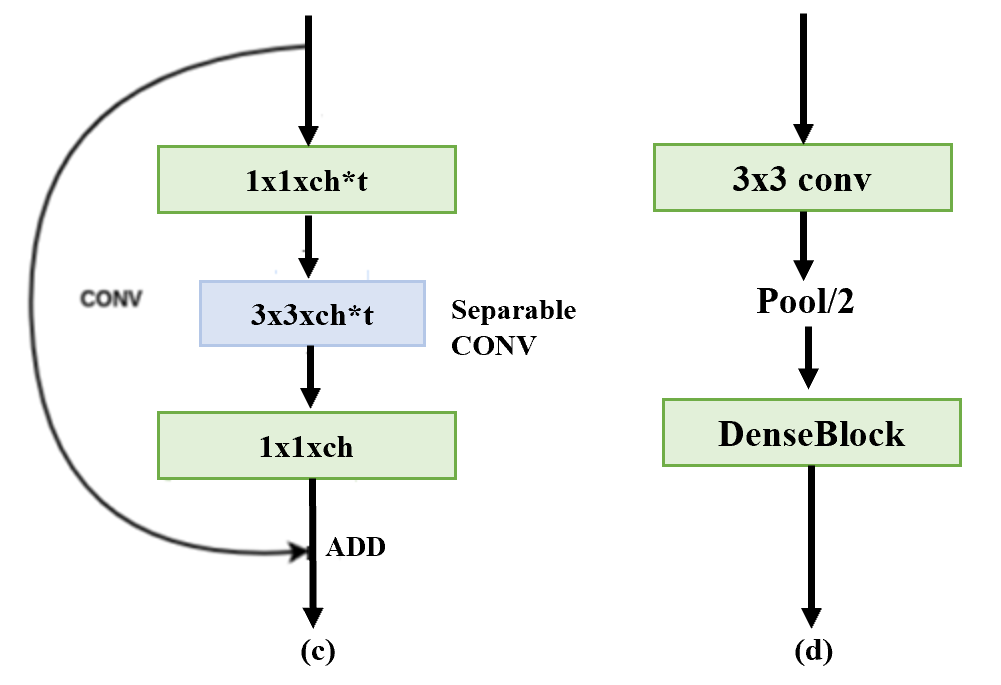}
\caption{Architecture of (a) VGG, (b) ResNet, (c) MobileNet, and (d) LightCNN.}
\label{figarc}
\end{figure}


\item \textbf{ResNet}~\cite{7780459} is a short form of residual networks based on the idea of "identity shortcut connection," where input features may skip certain layers. In this study, we used ResNet-50, which has $23.5M$ parameters.

\item \textbf{DenseNet}~\cite{huang2017densely} are inspired by residual networks, where all the previous layers' features are transferred to the current layer. Apart from tackling the vanishing gradients problem, this architecture also strengthens feature propagation and feature reuse while reducing the number of parameters required. In this study, we used DenseNet-121, which has $3.2M$ parameters.

\item \textbf{MobileNet}~\cite{howard2017mobilenets} is one of the most popular mobile-centric deep learning architectures, which is not only small in size but also computationally efficient while achieving high performance. The main idea of MobileNet is that instead of using regular $3\times3$ convolution filters, the operation is split into depth-wise separable $3\times3$ convolution filters followed by $1\times1$ convolutions. In our experiments, we used MobileNet-v2 with $0.5$x channels multiplier with an input size of $224\times 224$ for testing. This model is denoted as~\textit{MobileNet\_V2\_0.5\_224} in Table~\ref{tab:model_comp}.

\item \textbf{lightCNN}~\cite{DBLP:journals/corr/WuHS15}  This model heavily applies Max-Feature-Map (MFM) operation instead of ReLu activation. This acts as feature filter after each convolution layer. The operation takes two feature maps, eliminates the element-wise minimum, and returns element-wise maximum. By doing so across feature channels, only $50\%$ of the information-bearing nodes from each layer reach the next. Consequently, each layer is forced to preserve compact feature maps during training. The architecture is a stack of convolutional and MFM operations. For the purpose of this study, we used lightCNN consisting of $29$ layers and $11$M parameters and trained on ImageNet dataset from scratch.


\end{enumerate}



\begin{table}
\caption{Extracted feature size and number of parameters for each CNN model tested in this study. }
\begin{center}
\scalebox{0.99}{
\begin{tabular}{|l|c|c|}
\hline
\multicolumn{1}{|c|}{\textbf{Model}} & \textbf{Feature Size} & \textbf{Parameters}  \\ \hline
\textbf{VGG - 19}~\cite{Simonyan14c}                    & 4096                  & 140M                    \\ \hline
\textbf{ResNet - 50}~\cite{7780459}                 & 2048                  & 23.5M                  \\ \hline
\textbf{DenseNet - 121}~\cite{huang2017densely}              & 1024                  & 7M                        \\ \hline
\textbf{MobileNet\_v2\_0.5\_224}~\cite{howard2017mobilenets}      & 1280                  & 688K                       \\ \hline
\textbf{lightCNN}~\cite{DBLP:journals/corr/LiuWYLRS17}               & 512                  & 11M                          \\ \hline
\end{tabular}}
\label{tab:model_comp}
\end{center}
\end{table}

\subsection{End-to-End Convolutional Neural Network (CNN)}
For the sake of completeness and comparison, we also developed custom CNN trained from scratch for the performance evaluation.
Table~\ref{tab:cnn_arch} show the complete architecture of the proposed end-to-end CNN developed from scratch. The proposed CNN accepts input as $224\times 224$ image and consists of three convolutional layers followed by batch normalization and max pooling. These layers are followed by two fully connected layers of 200 channels and one channel for regression. The proposed CNN architecture is selected based on the performance obtained on the validation set. The model is trained using Adam optimizer~\cite{kingma2014adam} at a learning rate of $0.001$ using $150$ epochs and MAE as the loss function.

\begin{table}
\centering
\caption{Architecture of the custom end-to-end CNN for BMI prediction consisting of three convolutional blocks followed by two dense layers.}
\scalebox{0.9}{
\begin{tabular}{lll}
\textbf{Layer}     & \textbf{Output Shape}  & \textbf{\# Parameters} \\ \hline
conv1(3,3)    & (224, 224, 32) & 896 \\
batch normalization & (224, 224, 32) &     896 \\
max pooling(2,2) &  ( 112, 112, 32)  &    0 \\
conv2(3,3)     &      ( 112, 112, 64)   &   18496  \\

batch normalization & (112, 112, 64)   &   448  \\

max pooling(2,2) & (56, 56, 64)   &     0    \\

conv3(3,3)     &      (56, 56, 128)   &    73856   \\

batch normalization & (56, 56, 128)    &   224   \\

max pooling(2,2) & (28, 28, 128)    &   0     \\

flatten   &      (100352)        &   0      \\

dense      &       (200)        &       20070600   \\

dense         &      (1)          &       201 \\ \hline

\textbf{Total}               & 20,164,833            &                    \\ \hline
\end{tabular}}

\label{tab:cnn_arch}
\end{table}

\section{Experimental Validations}

\subsection{Dataset and Protocol}


\begin{itemize}

\item \textbf{VisualBMI dataset~\cite{Kocabey}:} This dataset comprises of total of $4206$ faces with corresponding gender and BMI information collected from the web. Of these, seven were in the under-weight range ($16$ $<$ BMI $\geq$ $18.5$), $680$ were normal ($18.5$ $<$ BMI $\geq$ $25$), $1151$ were overweight ($25$ $<$ BMI $\geq$ $30$), $941$ were moderately obese ($30$ $<$ BMI $\geq$ $35$), $681$ were severely obese ($35$ $<$ BMI $\geq$ $40$) and $746$ were very severely obese($40$ $<$ BMI). The subset of $2896$ images were used as the training set and the rest $1302$ images ($651$ male and $651$ female) as the test set. Training and testing subsets are selected to ensure equal number of male and females samples in the test set.

\item \textbf{VIP-Attributes~\cite{8546159}:}  Images in the VIP attribute dataset, are obtained in $2017$ from the WWW consisting of $513$ female and $513$ male subjects (mainly actors, singers and athletes). The images include the frontal pose of the subjects. Co-variates include illumination, expression, image quality, and resolution. Further challenging in this dataset are beautification (e.g., photoshop) of the images, as well as the presence of makeup, plastic surgery, beard, and mustache. The annotations related to the subjects' BMI were obtained from websites such as www.celebheights.com, www.howtallis.org, and celebsize.com. The subset of $726$ images was used as the training set and the rest $300$ samples as the test set ($150$ male and $150$ female).

\item \textbf{Bollywood dataset~\cite{bolly}:} The Bollywood data set was obtained from Github and had a total of $236$ images belonging to $231$ males and rest females. This is a small size dataset used for the sole purpose of evaluation of overall error rate.

\end{itemize}

\begin{figure} [!t]
\begin{center}
\label{images}
\includegraphics[scale=0.90]{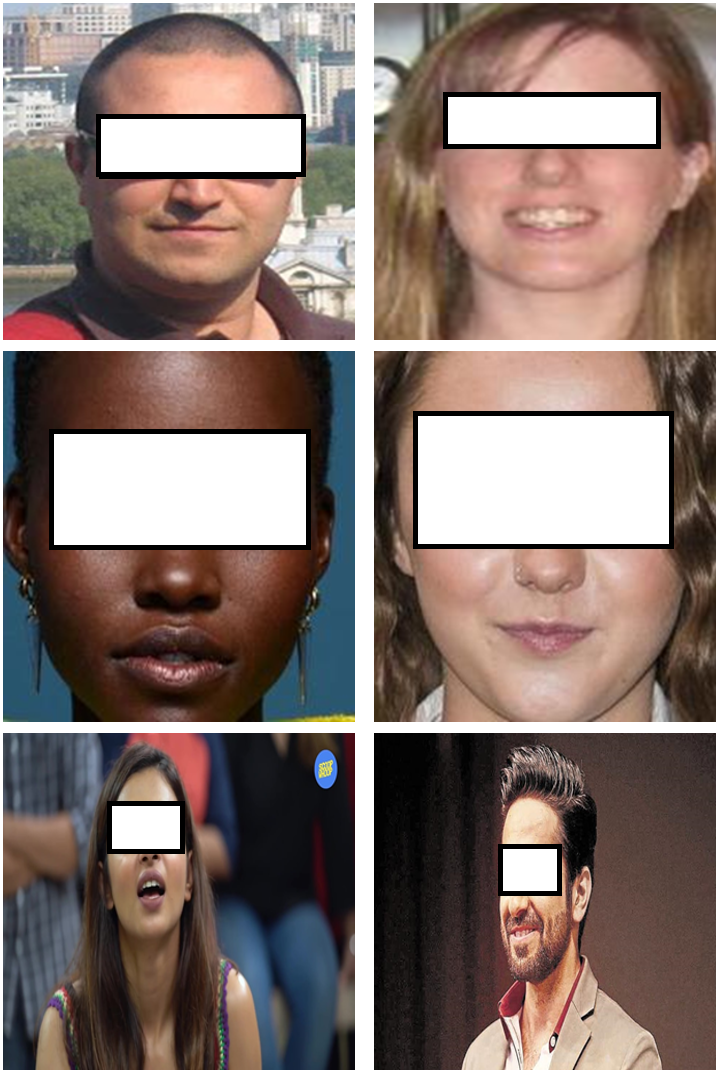}
\caption{Example of face images from VisualBMI~\cite{Kocabey}, VIP attribute~\cite{8546159} and bollywood datasets~\cite{bolly} row-wise. The covariates such as make-up, pose, lighting variations and uncontrolled background are available in these datasets.}
\end{center}
\end{figure}

\noindent Figure 1 shows the sample face images from VisualBMI~\cite{Kocabey}, VIP attribute~\cite{8546159} and Bollywood datasets~\cite{bolly} used in this study. The face detection was performed using Dlib library~\cite{dlib}, which is based on a Histogram of Oriented Gradients (HOG) used along with Support Vector Machine (SVM). Only eight face images from the VisualBMI dataset that obtained failure to enroll (FTE) error, caused due to failure in face detection, were discarded.
Mean absolute error (MAE) is used as a measure of the difference between the BMI ($\hat{BMI}$) inferred by the system and the ground truth ($BMI$) averaged over $n$ face images (eq.~\ref{eq:1}).

\begin{align}
MAE = \frac{\sum^{n}_{i=1}|\hat{BMI}_i-BMI_i|}{n}
\label{eq:1}
\end{align}

Using the training set of facial images (cropped face images obtained using face detection) from the VisualBMI and VIP-Attributes datasets, deep features were extracted from the pretrained models listed in Table~\ref{tab:model_comp}. The extracted deep features along with the Support Vector Regression (SVR) and Ridge Regression (RR) were used for training the BMI prediction model. The deep features extracted from the face images in the test set were used for testing the SVR and RR models using MAE. The Support Vector Regression (SVR) uses the same principles as the Support Vector Machine that is used for classification and outputs a real number. Ridge regression minimizes squared error while regularizing the norm of the weights as follows:
\begin{equation}
  J(w) = \lambda{w}^2 + \sum_i (w^T \hat{BMI}_i - {BMI}_i)^2.
\end{equation}

\subsection{Results}
Table~\ref{mae_mit} shows the MAE obtained for deep features extracted from VGG, ResNet, DenseNet, MobileNet, and lightCNN, used along with Support Vector Regression~(SVR) and Ridge Regression~(RR) for training and testing BMI prediction on VisualBMI dataset. It can be seen that all the CNNs obtained equivalent MAE in the range $[5.02, 5.87]$. No significant differences were noted across gender and regression methods (SVR and RR). DenseNet and ResNet obtained better performance overall when used with Ridge Regression. The study by Kocabey et al.~\cite{Kocabey} used the Pearson Correlation coefficient to evaluate the VGG model on the VisualBMI dataset. Therefore our results cannot be compared with this study~\cite{Kocabey}.

\begin{table}[t!]
\caption{MAE of deep features extracted from VGG, ResNet, DenseNet, MobileNet, and lightCNN, used along with support ector regression (SVR) and ridge regression (RR) for BMI inference on VisualBMI dataset.}
\begin{center}
\scalebox{0.99}{
\begin{tabular}{l|ll|ll|ll}
\multicolumn{1}{c}{} & \multicolumn{2}{c}{\textbf{MAE-Overall}} & \multicolumn{2}{c}{\textbf{MAE-Male}} & \multicolumn{2}{c}{\textbf{MAE-Female} }\\ \hline
     Model                     & RR          & SVR           & RR         & SVR          & RR          & SVR           \\ \hline
VGG-19~\cite{Simonyan14c}                    & 5.71            & 5.87          & 5.85          & 5.39         & 5.58           & 6.35          \\
ResNet-50~\cite{7780459}                    & 5.08            & 5.16          & 5.00             & 5.18         & 5.16           & 5.15          \\
DenseNet~\cite{huang2017densely}                  & 5.02            & 5.05          & 5.01            & 5.09         & 5.03           & 5.01          \\
MobileNet~\cite{howard2017mobilenets}                 & 5.56            & 5.39          & 5.78          & 5.23         & 6.44           & 6.12           \\
lightCNN~\cite{DBLP:journals/corr/LiuWYLRS17}               & 5.75            & 5.77          & 5.56          & 5.70          & 5.95           & 5.85          \\ \hline

\end{tabular}}
\end{center}
\label{mae_mit}
\end{table}

Table~\ref{mae_holly} shows the MAE obtained for deep features extracted from VGG, ResNet, DenseNet, MobileNet and lightCNN. These features are used along with support vector regression and ridge regression for training and testing BMI prediction on VIP Attribute dataset. It can be seen that all the CNNs obtained equivalent MAE in the range $[1.13, 2.57]$. ResNet and DenseNet obtained lower error when used along with Ridge Regression. No significant differences in error rates were noted across gender. For most of the cases, Ridge Regression obtained lower error rates than Support Vector Regression. Lower errors were obtained on the VIP Attribute dataset in comparison to the VisualBMI dataset. This could be due to low variance in the BMI annotation of the VIP Attribute dataset as it consists of facial images from Hollywood celebrities.

\begin{table}[t!]
\caption{MAE of deep features extracted from VGG, ResNet, DenseNet, MobileNet, and lightCNN used along with support vector regression (SVR) and ridge regression (RR) for BMI inference on VIP Attribute dataset.}
\begin{center}
\scalebox{0.99}{
\begin{tabular}{l|ll|ll|ll}
\multicolumn{1}{c}{} & \multicolumn{2}{c}{\textbf{MAE-Overall}} & \multicolumn{2}{c}{\textbf{MAE-Male}} & \multicolumn{2}{c}{\textbf{MAE-Female}} \\ \hline
     Model                     & RR           & SVR           & RR        & SVR          & RR          & SVR           \\ \hline
VGG-19~\cite{Simonyan14c}                    & 2.44             &  2.57          & 2.39         & 2.42         & 2.49            & 2.73          \\
ResNet-50~\cite{7780459}                   & 1.13            & 1.74          & 1.10             & 1.73         & 1.17             & 1.75          \\
DenseNet~\cite{huang2017densely}                  & 1.14            & 1.33          & 1.11             & 1.24         & 1.17             & 1.43          \\
MobileNet~\cite{howard2017mobilenets}                  & 2.24            & 2.21          & 2.16          & 2.25         & 2.32          & 2.18           \\
lightCNN~\cite{DBLP:journals/corr/LiuWYLRS17}               & 2.42             & 2.32          & 2.75          & 2.17         & 2.10           &2.47          \\ \hline

\end{tabular}}
\label{mae_holly}
\end{center}
\end{table}

Table~\ref{mae_bolly} shows the MAE obtained for deep features extracted from VGG, ResNet, DenseNet, MobileNet and
lightCNN for BMI prediction on Bollywood dataset. It can be seen that all the CNNs obtained equivalent MAE in the range $[1.04, 2.98]$. ResNet and DenseNet, when used along with Ridge Regression, obtained the lowest error over other CNNs.

Table~\ref{custom} shows the MAE obtained for end-to-end CNN for BMI prediction on Visual BMI, VIP Attributes, and Bollywood datasets. It can be seen that MAE was obtained in the range $[4.12, 6.65]$. The performance of the custom CNN is lower in comparison to those obtained using pretrained CNNs for all the datasets~\cite{8546159}. This is obviously due to limited training dataset causing over-fitting.  Even for custom CNN, no significant differences have been noted across male and female.

\begin{table}[t!]
\caption{MAE of deep features extracted from VGG, ResNet, DenseNet, MobileNet,and lightCNN, used along with support vector regression (SVR) and ridge regression (RR) for BMI inference on bollywood dataset.}
\begin{center}
\begin{tabular}{ccc}
      & \multicolumn{2}{c}{\textbf{MAE-Overall}} \\ \hline
   Model        & RR           & SVR           \\ \hline
VGG-19~\cite{Simonyan14c}     & 1.49            & 1.99          \\
ResNet~\cite{7780459}     & 1.04           & 1.85          \\
DenseNet~\cite{huang2017densely}     & 1.39            & 1.65          \\
MobileNet~\cite{howard2017mobilenets}   & 2.10            & 2.98          \\
lightCNN~\cite{DBLP:journals/corr/LiuWYLRS17} & 1.90            & 1.95         \\ \hline
\end{tabular}
\end{center}
\label{mae_bolly}
\end{table}

\begin{table}[t!]
\caption{MAE of custom CNN for BMI inference on Visual BMI, VIP Attributes and Bollywood datasets.}
\begin{center}
\scalebox{0.99}{
\begin{tabular}{cccc} \hline
\textbf{Dataset}        & \textbf{MAE-Overall} & \textbf{MAE- Male} & \textbf{MAE-Female} \\ \hline
VisualBMI~\cite{Kocabey}      & 6.48        & 6.65      & 6.32       \\
VIP Attributes~\cite{8546159} & 4.39        & 4.66      & 4.12       \\
Bollywood~\cite{bolly}      & 4.33        & N/A       & N/A       \\ \hline
\end{tabular}
}
\end{center}
\label{custom}
\end{table}

For all the experiments, ResNet and DenseNet consistently outperformed other networks. This could be due to the advantages of ResNet and DenseNet such as they alleviate the vanishing-gradient problem, strengthen feature propagation, and encourage feature reuse.

\section{Conclusion and future work}
In this paper, we investigated deep learning-based methods for BMI inference from facial images.
Experimental investigations on three publicly available facial image datasets obtain overall MAE in the range $[1.04, 6.48]$. The results varied across datasets due to variance in the BMI annotation and sample size difference.
No significant differences in error were noted across gender. DenseNet and ResNet obtained superior performance over other nets.
As a part of future work, an extended list of CNNs will be compared on the large face dataset assembled using mobile devices across different ethnicities and age groups for statistical validation of the reported results. Compact size custom CNNs with low latency will be investigated for on-device deployment in smartphones. Further, the impact of facial covariates on BMI prediction will be statistically quantified.


\small{
\bibliographystyle{IEEE}
\bibliography{refs}}

\begin{thebibliography}{10}

\bibitem{6248021}
W.~J. {Scheirer}, N.~{Kumar}, P.~N. {Belhumeur}, and T.~E. {Boult},
\newblock ``Multi-attribute spaces: Calibration for attribute fusion and
  similarity search,''
\newblock in {\em 2012 IEEE Conference on Computer Vision and Pattern
  Recognition}, June 2012, pp. 2933--2940.

\bibitem{selfie}
A.~Rattani, R.~Derakhshani, and A.~Ross,
\newblock {\em Selfie Biometrics},
\newblock Springer, 2019.

\bibitem{bio_iet}
A.~Rattani, N.~Reddy, and R.~Derakhshani,
\newblock ``Convolutional neural networks for gender prediction from
  smartphone-based ocular images,''
\newblock {\em IET Biometrics}, vol. 7, pp. 423--430, 2018.

\bibitem{8272766}
A.~{Rattani}, N.~{Reddy}, and R.~{Derakhshani},
\newblock ``Convolutional neural network for age classification from
  smart-phone based ocular images,''
\newblock in {\em IEEE International Joint Conference on Biometrics}, Denver,
  CO, 2017, pp. 756--761.

\bibitem{jain04}
A.K. Jain, S.~C Dass, and K.~Nandakumar,
\newblock ``Soft biometric traits for personal recognition systems,''
\newblock in {\em International Conference on Biometric Authentication}, 2004,
  p. 731–738.

\bibitem{Wen13}
L.~Wen and G.D. Guo,
\newblock ``{A computational approach to body mass index prediction from face
  images},''
\newblock {\em Image and Vision Computing}, vol. 31, no. 5, pp. 392–400,
  2013.

\bibitem{8546159}
A.~{Dantcheva}, F.~{Bremond}, and P.~{Bilinski},
\newblock ``Show me your face and i will tell you your height, weight and body
  mass index,''
\newblock in {\em 2018 24th International Conference on Pattern Recognition
  (ICPR)}, Aug 2018, pp. 3555--3560.

\bibitem{Kocabey}
E.~Kocabey, M.~Camurcu, F.~Ofli, Y.~Aytar, J.~Marin, A.~Tor-ralba, and
  I.~Weber,
\newblock ``Face-to-bmi: Using computer vision toinfer body mass index on
  social media,''
\newblock in {\em arXiv:1703.03156}, 2017.

\bibitem{hales}
C.M Hales, M.D Carroll, C.D Fryar, and C.L. Ogden,
\newblock ``Prevalence of obesity among adults and youth: United states,
  2015-2016,''
\newblock {\em NCHS Data Brief}, vol. 288, pp. 1--8, 2017.

\bibitem{Wang}
Y.~C Wang, J.~Pamplin, M.W Long, Z.J Ward, S.~L Gortmaker, and T.~Andreyeva,
\newblock ``Severe obesity in adults cost state medicaid programs nearly dollar
  8 billion in 2013,''
\newblock {\em Health Aff}, pp. 1923--1931, 2015.

\bibitem{Mann2020}
Mann DM, Chen J, Chunara R, and Testa PA,
\newblock ``Covid-19 transforms health care through telemedicine: evidence from
  the field,''
\newblock {\em J Am Med Inform Assoc.}

\bibitem{Simonyan14c}
K.~Simonyan and A.~Zisserman,
\newblock ``Very deep convolutional networks for large-scale image
  recognition,''
\newblock {\em CoRR}, vol. abs/1409.1556, 2014.

\bibitem{7780459}
K.~{He}, X.~{Zhang}, S.~{Ren}, and J.~{Sun},
\newblock ``Deep residual learning for image recognition,''
\newblock in {\em IEEE Conference on Computer Vision and Pattern Recognition
  (CVPR)}, June 2016, pp. 770--778.

\bibitem{huang2017densely}
G.~{Huang}, Z.~{Liu}, L.~v.~d. {Maaten}, and K.~Q. {Weinberger},
\newblock ``Densely connected convolutional networks,''
\newblock in {\em IEEE Conference on Computer Vision and Pattern Recognition
  (CVPR)}, July 2017, pp. 2261--2269.

\bibitem{howard2017mobilenets}
A.~G. Howard, M.~Zhu, B.~Chen, D.~Kalenichenko, W.~Wang, T.~Weyand,
  M.~Andreetto, and H.~Adam,
\newblock ``Mobilenets: Efficient convolutional neural networks for mobile
  vision applications,''
\newblock {\em arXiv preprint arXiv:1704.04861}, 2017.

\bibitem{DBLP:journals/corr/WuHS15}
Xiang Wu, Ran He, and Zhenan Sun,
\newblock ``A lightened {CNN} for deep face representation,''
\newblock {\em CoRR}, vol. abs/1511.02683, 2015.

\bibitem{DBLP:journals/corr/LiuWYLRS17}
W.~Liu, Y.~Wen, Z.~Yu, M.~Li, B.~Raj, and L.~Song,
\newblock ``Sphereface: Deep hypersphere embedding for face recognition,''
\newblock {\em CoRR}, 2017.

\bibitem{kingma2014adam}
Diederik~P. Kingma and Jimmy Ba,
\newblock ``Adam: A method for stochastic optimization,''
\newblock {\em CoRR}, vol. abs/1412.6980, 2014.

\bibitem{bolly}
``Bollywood dataset,''
  \url{https://github.com/abhaymise/Face-to-height-weight-BMI-estimation-/blob/master/BMI\%20data\%20-\%20Sheet1.csv}.

\bibitem{dlib}
Chunara R Testa~PA Mann~DM, Chen~J,
\newblock ``Dlib-ml: A machine learning toolkit,''
\newblock {\em Journal of Machine Learning Research}, vol. 10, pp. 1755--1758,
  2009.

\end{thebibliography}

\end{document}